\RequirePackage[T1]{fontenc}
\documentclass[conference,10pt,letterpaper]{IEEEtran}
\newif\ifIEEEFormat
\IEEEFormattrue
\usepackage{amsmath}
\ifIEEEFormat
  \usepackage{amssymb}
  \usepackage{cite}
  \usepackage[hidelinks]{hyperref}
\else
  \setcopyright{none}
  \acmConference{}{}{}
  \acmBooktitle{}
  \acmDOI{}
  \acmISBN{}
  \acmYear{}
  \copyrightyear{}
  \acmPrice{}
\fi
\usepackage{graphicx}
\usepackage{booktabs}
\usepackage{array}
\usepackage{xcolor}
\hypersetup{hidelinks}

\newcommand{\ours}{EDAGoat}

\newcommand{\resultpending}[1]{\textcolor{red!65!black}{[TBD\_RESULT: \nolinkurl{#1}]}}
\newcommand{\resultcell}[1]{\textcolor{red!65!black}{\textsf{TBD}}}

  \title{EDAGoat: An EDA-Aware Agent\\for Project-Level GPU Optimization}
  \author{\IEEEauthorblockN{Chengxi Li, Liang Xiao, Yutao Wang, Wenhao Lin, Xinzhuang Xiong, Han Zhao and Evangeline Young}
    \IEEEauthorblockA{Department of Computer Science and Engineering, The Chinese University of Hong Kong, Hong Kong SAR, China\\
    \{cxli23, lxiao23, ytwang1, whlin23, xzxiong26, hzhao, fyyoung\}@cse.cuhk.edu.hk}}

\begin{document}
\maketitle
\begin{abstract}
GPU acceleration has transformed electronic design automation (EDA), yet further improving expert-engineered GPU tools still demands substantial manual effort. Automating this work calls for coding agents that can trace bottlenecks across host and device code while preserving correctness and design quality. We introduce \ours, to our knowledge the first EDA-aware GPU optimization agent for this task. Its optimization enhancement (OE) mode guides edits with four cross-project GPU optimization patterns. Its mandatory validation (MV) mode requires the agent to test the edited program against the EDA constraints its changes could affect.
Our benchmark covers five GPU EDA projects across four task families and 55 circuit inputs, measuring complete-solve runtime and verifying the resulting designs.
Each configuration runs three independent searches per project. A search qualifies only if its solver passes correctness and design-quality checks on all public and hidden inputs. With DeepSeek, generic prompting qualifies in 7 of 15 searches and OE in 6 of 15, despite a higher best speedup on four projects. Adding MV raises this to 13 of 15 and, on every project, exceeds both generic baselines in mean and best speedup and OE in mean speedup; the geometric mean of per-project mean speedups rises from $1.11\times$ ($1.36\times$ with OE) to $1.53\times$.
With GPT-6 Astra, \ours\ qualifies in all 15 searches and improves both speedup metrics over generic prompting on every project, lifting the same geometric mean from $1.26\times$ to $1.57\times$.
These results show that agents can use EDA knowledge to find further speedups in expert-tuned tools while preserving design correctness and quality, establishing EDA-aware GPU optimization as a new direction for agent research.
\end{abstract}

% Author metadata for the local arXiv edition.
\hypersetup{pdfauthor={Chengxi Li, Liang Xiao, Yutao Wang, Wenhao Lin, Xinzhuang Xiong, Han Zhao, Evangeline Young},
  pdftitle={EDAGoat: An EDA-Aware Agent for Project-Level GPU Optimization},
  pdfsubject={},pdfkeywords={}}

\section{Introduction}
\label{sec:intro}
GPU acceleration shortens some of the most expensive stages of electronic design automation (EDA), from logic synthesis and placement to routing and timing analysis~\cite{culs,xplace,instantgr,gcstimer}. Engineers repeat these stages as a circuit evolves, so faster GPU implementations let them evaluate more design alternatives within the same schedule. DREAMPlace, for example, reports roughly 40-fold faster global placement than a multithreaded CPU placer at comparable quality~\cite{dreamplace}. Further gains demand expert time to trace bottlenecks across host and device code and machine time to build and validate each revision. Coding agents already perform parts of this cycle. By editing and profiling existing CUDA kernels from SGLang, an LLM serving framework, a multi-agent system achieved a mean speedup of 1.32$\times$~\cite{astra}. \emph{How much further can a coding agent accelerate the GPU EDA tools that experts have already built?}

\begin{figure*}[t]
\centering
\includegraphics[width=.91\textwidth]{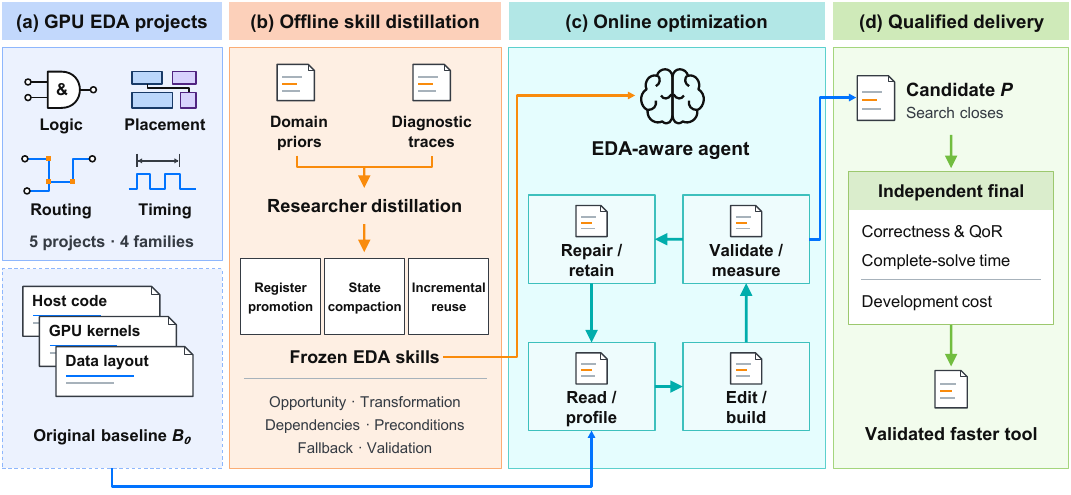}
\caption{\textbf{How EDAGoat optimizes a GPU EDA project.} (a) The agent starts from the original baseline $B_0$ and (b) receives a frozen library of cross-project patterns and validation templates. (c) It edits and tests the project; mandatory validation requires evidence from the current build. (d) An independent evaluator checks the nominated candidate $P$ and times it against $B_0$. The symbol \& denotes AND.}
\label{fig:method}
\end{figure*}

Existing work approaches this question from two directions. EDA agents automate design flows and edit the source of tools such as OpenROAD and ABC to improve design quality~\cite{chateda,audopeda,selfabc}. General GPU optimization agents use hardware feedback to revise kernel implementations~\cite{cudaforge,hiera}. These advances leave open how an agent can accelerate a complete GPU EDA solver while preserving the correctness and quality of the designs it produces. We introduce \ours, an EDA-aware GPU optimization agent for this project-level task. As Fig.~\ref{fig:method} shows, the agent uses a frozen library of optimization patterns and validation templates to develop an optimized implementation and submits one candidate for independent evaluation. Its workflow targets two capabilities: finding faster code and delivering a qualified program.

The first challenge is finding redundant or unevenly distributed GPU work in irregular circuit computations. Much of the computation in logic synthesis, routing, and placement is irregular~\cite{culs,instantgr,xplace}: its cost varies with the size of a logic cone, the pin count of a net, or the footprint of a cell. In placement, assigning one thread per cell can leave one thread processing a large cell's density bins long after its neighbors finish. Sharing that work across lanes requires covering each bin once, combining partial forces, and synchronizing even a partially filled warp.

The second challenge is delivering a qualified program: thread-level changes must satisfy several design constraints at once.
A faster placer must preserve fixed cells, produce a legal placement, and meet a wirelength bound. A logic rewrite must preserve Boolean equivalence while meeting bounds on both AND count and depth. An agent must therefore check correctness and quality of results (QoR) on the final design and measure speed across the complete solve. 
In our global routing benchmark, for instance, evaluating one candidate across ten large designs requires twelve end-to-end routing solves per design (one warm-up pair and five timed baseline--candidate pairs) plus a contest verifier run on every output route to check net coverage, 3D tree connectivity, wirelength, and overflow. This takes over three hours, so an agent cannot afford full evaluation after every edit and must rely on targeted checks.

\emph{Qualified delivery is the foundation of useful acceleration.} If a program fails even one required check, its speedup cannot count toward the reported result, however fast it runs on the inputs where it succeeds. Improving delivery rate and improving speedup are therefore equally important. 
Delivery is hardest when edits change how solver components share state: the agent must identify which constraints they could violate and run checks that expose those failures before submission. Passing ordinary development inputs does not guarantee that a new representation, synchronization scheme, or memory requirement supports the full legal input range.

To address both challenges, EDAGoat combines reusable domain optimization knowledge with enforced verification through two modes in Fig.~\ref{fig:method}(c). First, domain optimization principles reveal which work can be avoided or redistributed and when doing so is safe: for instance, a global router can reuse cumulative costs along unaffected tracks, provided it refreshes every track touched by a rerouted net before the next routing decision. \textbf{Optimization enhancement (OE) mode} provides a frozen library of cross-project optimization patterns and validation templates distilled from such transformations, dependencies, and validation logic. The agent must still locate the relevant modules in the original project, design the concrete GPU thread and memory mappings, and test its edits. Second, because validation guidance alone does not ensure that an agent runs the relevant checks during complex edits (Section~\ref{sec:eval}), \textbf{mandatory validation (MV) mode} pairs validation templates with a workflow controller. Before an agent can advance a build to performance evaluation or final nomination, the controller requires execution evidence from targeted checks on the current implementation and forces the agent to repair any exposed defects.

We evaluate EDAGoat with two base models, GPT-6 Astra (hereafter Astra; unrelated to the kernel optimizer in~\cite{astra}) and DeepSeek-v4.1-Flash, on five expert-engineered GPU EDA tools and 55 inputs. With Astra, EDAGoat keeps qualified delivery at 15/15 while improving mean and best complete-solve speedup on every project, reaching up to $3.906\times$ on logic synthesis. With DeepSeek, EDAGoat produces the fastest qualified solver on every project; MV also raises qualified delivery from $7/15$ for generic prompting (and $6/15$ for OE alone) to $13/15$ and achieves higher mean speedup than generic prompting and OE on all five projects.

Our main contributions are as follows:
\begin{itemize}
    \item We propose EDAGoat, a project-level agent workflow with two modes: OE supplies four cross-project GPU optimization patterns paired with validation templates, and MV adds a rule-based controller that blocks performance evaluation and nomination until targeted checks pass on the current build.
    \item We build a benchmark of five expert-engineered GPU EDA solvers spanning logic synthesis, placement, global routing, and static timing analysis, with 55 public and hidden inputs and per-project equivalence, legality, and QoR contracts that every delivered program must satisfy.
    \item We show that generic agents fail mainly through GPU write races and changes to the required computation, and that for a less reliable model, pattern guidance alone raises best speedup but not delivery, which requires enforced validation.
\end{itemize}

% Input early so Table~\ref{tab:tasks} floats to the page where Section~\ref{sec:benchmark} starts.
\begin{table*}[t]
\centering
\caption{Benchmark tasks, timed work, and qualification criteria. Inputs are total (public/hidden). GP, LG, and DPL denote global placement, legalization, and detailed placement. Each task fixes source, algorithm, precision, and work requirements. Total and OverflowScore follow the ISPD 2024 contest metrics.}
\label{tab:tasks}
\footnotesize
\setlength{\tabcolsep}{2pt}
\renewcommand{\arraystretch}{0.92}
\begin{tabular*}{\textwidth}{@{\extracolsep{\fill}}llc >{\raggedright\arraybackslash}p{4.65cm}!{\hspace{8pt}} >{\raggedright\arraybackslash}p{7.25cm}@{\hspace{2pt}}}
\toprule
Project & Family & Inputs & Primary timed computation & Required output and quality checks \\
\midrule
CULS & Logic synthesis & 10 (4/6) & Read, build, rewrite, reconstruct host AIG & ABC combinational equivalence; AND count and depth each at or below $B_0$ \\
InstantGR (IG) & Routing & 10 (4/6) & Complete \texttt{route()} interval & Original-net coverage, connectivity, geometry; \mbox{$\text{Total}(P) \le 1.01\,\text{Total}(B_0)$} and \mbox{$\text{OverflowScore}(P) \le 1.0025\,\text{OverflowScore}(B_0)$} \\
DREAMPlace (DP) & Placement & 13 (4/9) & Database preparation, GP/LG/DPL, original-object output & Original objects, fixed coordinates/tags, legal placement; \mbox{$\mathrm{HPWL}(P) \leq 1.005\,\mathrm{HPWL}(B_0)$} \\
Xplace (XP) & Placement & 12 (4/8) & \mbox{Parser/database}, \mbox{GP/LG/DPL}, output-unit records & Original objects, fixed coordinates/tags, full legality; \mbox{$\mathrm{HPWL}(P) \leq 1.005\,\mathrm{HPWL}(B_0)$}; fixed deterministic flow \\
GCS-Timer (GCS) & Timing & 10 (4/6) & Library/circuit preparation, default graph-based analysis, complete host results & Required fields/identities, finite values; reference error \mbox{$\leq 10^{-4}\,\mathrm{ps}+10^{-9}\lvert\mathrm{ref}\rvert$} \\
\bottomrule
\end{tabular*}
\end{table*}

\section{Related Work}
\label{sec:related}
\textbf{From flow automation to solver evolution.}
EDA agents script and tune flows~\cite{chateda,jarvis,mcp4eda,autoeda,crop,alpharoute} or generate and repair hardware designs~\cite{rtlfixer,verilogcoder,mage,analogcoder,horizon,phoenixbench}. AuDoPEDA follows OpenROAD's repository graph to locate edits and evaluates patches by wirelength, timing, and power~\cite{audopeda}. Self-Evolved ABC rebuilds the synthesis binary and checks Boolean equivalence before comparing area and delay~\cite{selfabc}. Trace2Skill uses verifier diagnostics to select revised instructions for the next Verilog attempt~\cite{trace2skill}. GR-Evolve specializes a router per design, and its selection can favor shorter wires even if the router slows down~\cite{grevolve}. SATLUTION optimizes whole SAT solvers for speed, validating answers and unsatisfiability proofs before comparing runtimes~\cite{satlution}.

\textbf{GPU kernel optimization.}
KernelBench counts generated implementations that match a PyTorch reference and exceed a speedup threshold~\cite{kernelbench}. Wei et al.~\cite{astra} start from SGLang CUDA kernels and assign planning, editing, testing, and profiling to cooperating agents. CudaForge's judge returns correctness results and Nsight Compute metrics to its coder~\cite{cudaforge}. HIERA uses each workload's contract to choose among PyTorch operators, library calls, and custom CUDA~\cite{hiera}. KernelClaw remembers tuning decisions~\cite{kernelclaw}, while PCPO learns from populations of generated candidates~\cite{pcpo}. Repository-level benchmarks such as GSO and SWE-Perf validate speedups with unit tests~\cite{gso,sweperf}, which cannot capture the equivalence, legality, and QoR contracts of an EDA solver.

\textbf{Our position.}
In CULS, Xplace, InstantGR, GCS-Timer, and DREAMPlace~\cite{culs,xplace,instantgr,gcstimer,dreamplace}, host code and GPU kernels share irregular circuit data. A routing kernel can pass its first test yet return stale costs after a net moves. Conversely, different routes can satisfy the same connectivity and QoR checks, so a candidate cannot be validated by matching $B_0$'s output bit for bit. \ours\ targets this project-level setting, combining cross-project optimization patterns with controller-enforced validation and timing each qualified solver end to end.

\section{Background and Problem Formulation}
\label{sec:problem}
Given an original GPU EDA project $B_0$ (Fig.~\ref{fig:method}(a)), the agent submits one modified program $P$. Each task fixes the editable source, build recipe, development inputs, and budget. To qualify, $P$ must build, perform all required contract stages, and pass all correctness and QoR checks.

Local edits can change global design decisions. An and-inverter graph (AIG) rewrite changes a Boolean network, a placer moves cells that must later be legalized, and routing and timing propagate state across a design. These dependencies make the complete solve the unit of evaluation.

Let $q(P)=1$ for a qualified program and $q(P)=0$ otherwise. Delivery rate is the fraction of assigned searches with \mbox{$q(P)=1$.}\enspace Performance is the complete-solve speedup over $B_0$. A failed search falls back to $B_0$ and scores $1\times$. A high best-case speedup can therefore coexist with frequent delivery failures. We report both the arithmetic mean over assigned searches and the best result.

\section{Benchmark Construction and Protocol}
Our benchmark contains 55 inputs for five GPU EDA solvers: CULS, InstantGR (IG), DREAMPlace (DP), Xplace (XP), and GCS-Timer (GCS)~\cite{culs,instantgr,dreamplace,xplace,gcstimer} (Table~\ref{tab:tasks}). Each task fixes its inputs, qualification checks, and timing boundary.
\label{sec:benchmark}
\subsection{Benchmark Projects and Timing Boundaries}

CULS uses ten and-inverter graphs from the EPFL combinational benchmark suite~\cite{epfl}. 
IG uses five ISPD 2024 routing inputs~\cite{ispd2024}, three ISPD 2025 designs converted to routing-only inputs, and two placed designs from RosettaStone~\cite{rosettastone}, timing the complete \texttt{route()} solver stage as in~\cite{instantgr,ispd2024}. 
DP and XP use Bookshelf designs from the ISPD 2005 and 2006 placement contests~\cite{ispd2005,ispd2006}; XP excludes one design on which $B_0$ itself fails. GCS uses four EPFL-derived timing designs distributed with GCS-Timer and six additional EPFL circuits for which we extracted parasitics after placement and routing in ASAP7~\cite{asap7}. All agents develop on the same four small-to-medium public inputs per project; the rest, including the largest circuits, are hidden.

\subsection{Qualification Before Performance Acceptance}
An automated evaluation harness and a configuration-blind manual audit of each source diff check each nominee for omitted contract work. The harness checks every output against Table~\ref{tab:tasks}. For example, CULS must preserve Boolean equivalence, checked with ABC~\cite{abc}, without exceeding $B_0$ in AND count or depth; a placer must meet legality and half-perimeter wirelength (HPWL) bounds. QoR tolerances, fixed before any search, absorb $B_0$'s run-to-run variation from parallel accumulation order and tie-breaking.

During development, agents can test public inputs and construct their own. A hidden input can expose a defect that survived all development checks.

\subsection{Measuring the Final Candidate's Speedup}
For each paired run of $B_0$ and the fixed nominee $P$, speedup is the ratio of original to candidate solve time, $t_{B_0}/t_P$. A qualified candidate's reported speedup is the geometric mean across repetitions and inputs, with equal weight per input. A qualified slowdown retains its measured score, even when that score falls below the $1\times$ fallback.

\section{The EDAGoat Optimization Workflow}
\label{sec:method}
Figure~\ref{fig:method} follows the agent from the original project and frozen skill library to one independently evaluated build. Each skill pairs an optimization pattern with a validation template. Mandatory validation (MV) mode retains optimization enhancement (OE) and adds controller-enforced checks.

\subsection{Starting a Search from the Original Project}
Each search starts from its own copy of $B_0$ in a CLI harness with tools to edit, build, run, and profile. All agents receive the same task contract (Table~\ref{tab:tasks}); \ours\ agents also get the frozen library as read-only files. A deterministic rule-based controller, not an LLM judge, records the source and build hash of each candidate and charges builds, profiling, and tests to the search budget.

\subsection{Cross-Project Optimization Patterns and Validation Templates}
\label{sec:patterns}
We distill general GPU acceleration priors and early diagnostic traces on public inputs into four \emph{cross-project optimization patterns} and a set of \emph{generalizable validation templates} (Fig.~\ref{fig:method}(b)). We freeze the same library for all five projects before evaluation. It contains no project names, file paths, function signatures, source patches, or descriptions of a particular codebase. The agent inspects the solver, checks a pattern's preconditions, and chooses the representation and thread mapping. The agents produced the examples below.

\textbf{Bounded-state register specialization.} When local state has a known bound, the agent can replace dynamic indexing with fixed register slots or a packed integer. In CULS it kept up to four cut leaves in registers in truth-variable order. In XP it packed the bid and tie-breaking bidder ID into one integer whose ordering preserves the auction winner.

\textbf{Dependency-tracked incremental reuse.} The agent traces reads and writes to determine which state a local change invalidates. It reused cumulative cost prefixes on unaffected IG tracks, logic-level boundaries for narrow CULS levels, and buffers across DP calls after checking which outputs remained live. In each case, the next consumer determines when an invalidated value must be refreshed.

\textbf{Competitive-selection state compaction.} When only the winner survives a bidding round, one record per item can replace a dense bid matrix. The agent applied this pattern in both DP and XP, preserving bid order and deterministic tie-breaking in each placer.

\textbf{Warp/block-cooperative subproblem partitioning.} The agent assigns an irregular object or dense submatrix to cooperating lanes that share on-chip state. In DP it split force accumulation across four lanes for cells spanning more than 64 bins. In GCS it mapped independent RC or stage matrix inversions to blocks, with warps sharing pivot data and broadcasting elimination coefficients. Participating lanes must see the pivot row before updating dependent rows. Matrices that exceed on-chip capacity take the original path.

\textbf{Validation templates} specify how to test an invariant against a trusted reference. For cache reuse, a test populates the cache, changes one dependency, and compares the next output with a fresh computation. The template applies to routing costs and placement density maps alike. The agent writes the native driver, chooses a legal input that exercises the edit, and checks the affected outputs.

\subsection{The Optimization Loop and Its Validation Requirements}
\label{sec:modes}
In Fig.~\ref{fig:method}(c), OE lets the agent decide what to test and retain. MV strengthens two steps: \emph{Validate/measure} requires evidence from the current build, and \emph{Repair/retain} uses that evidence to decide whether the build can advance.

\subsubsection{Optimization Enhancement Mode}
\begin{figure}[t]
\centering
\includegraphics[width=.91\columnwidth]{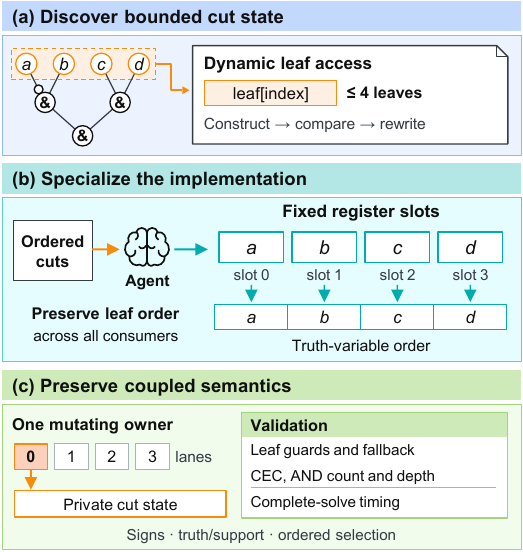}
\caption{\textbf{CULS ordered-cut optimization.} (a) A bounded cut (\& is AND; circles are inversions). (b) Register slots keep truth-variable order. (c) Ownership and combinational equivalence checking (CEC) guard the rewrite.}
\label{fig:cut-skill}
\end{figure}

\textbf{Read and profile.} The agent uses a skill to identify a suspected bottleneck and inspects the relevant stage, profiling it when needed. In CULS, the rewrite kernel indexes cut leaves dynamically even though a cut has at most four leaves (Fig.~\ref{fig:cut-skill}(a)). The agent checks the bound at the callers before changing the representation.

\textbf{Edit and build.} The agent traces each cut from construction through comparison to truth-table evaluation. Four explicit register slots replace dynamic indexing, but every writer and reader must preserve leaf position and sign (Fig.~\ref{fig:cut-skill}(b)). The mapping in panel~(c) gives one lane ownership of the mutable cut set; disjoint slot ownership is also valid.

\textbf{Validate and measure.} On public circuits, the agent checks combinational equivalence and the bounds on AND count and depth. It may add its own diagnostics and times the complete rewrite to see whether faster leaf access speeds up the solver.

\textbf{Repair or retain.} Failed checks send the agent back to the affected code. In one CULS search, Astra compared further loop unrolling with the retained register-cut build, swapped their timing order to rule out run-order bias, and discarded the extra unrolling. The next edit started from the faster retained build.

\subsubsection{Mandatory Validation Mode}
\label{sec:mandatory}

\textbf{Validate and measure.} The agent checks each property its edit could violate, comparing against $B_0$ or an independent reference. The cut in Fig.~\ref{fig:cut-skill}(a) computes $(\neg a \land b) \land (c \land d)$. On input $(0,1,1,1)$ it returns~1; with $a$ and $b$ swapped, it returns~0. Comparing all 16 assignments checks the complete truth table. The controller marks the current build \texttt{READY} only if the recorded tests ran successfully on its source tree and build hash, cover each changed component as the project policy requires, and leave no earlier failure unresolved; otherwise it returns \texttt{INCOMPLETE}.

Each project policy is derived from Table~\ref{tab:tasks} and frozen before evaluation. \emph{CULS} gates nomination on equivalence, AND count, and depth for all four public circuits. The agent chooses targeted checks for edits such as cut reordering. \emph{IG} requires a \texttt{READY} assessment tied to successful current-build executions before profiling, evaluation, comparison, or nomination. A cached routing prefix needs a legal update sequence and a reference that exposes stale costs. Storage edits need a peak-memory bound and a fallback for larger inputs.

\emph{DP} checks CUDA density output against a reference on clipped bin overlaps. Buffer reuse requires successive calls with changed input and checks of outputs that must survive the second call. \emph{XP} compares parsed database fields with $B_0$. A reorder edit requires cases that distinguish duplicate-pin order from the limit on collected pin occurrences. Before measurement or nomination, both project policies require tests of the changed component on the current build. A test must compare all supplied outputs and reject a deliberately incorrect result; the agent must account for earlier failures before advancing.

\emph{GCS} links each changed timing function to the test entry points that execute it. Tests follow the producer and consumer of each changed matrix, covering its orientation, both signal transitions, and the timing fields that depend on them. The controller withholds performance timings from correctness and repair feedback until the build is \texttt{READY}.

\textbf{Repair or retain.} An \texttt{INCOMPLETE} assessment identifies missing evidence or a failed relation. The agent can repair the implementation, fix a faulty test driver, or revert to an earlier validated build. After a source edit, the agent must rerun the affected checks on the revised build. A later failure reopens validation. Complete-solve timing then guides retention or nomination. Diagnostics, builds, and repairs count against the search budget.

\subsection{Final Candidate Submission and Evaluation}
The agent uses development checks and timing to select one build $P$ as its final candidate. Nomination closes the search and fixes the build for independent evaluation (Fig.~\ref{fig:method}(d)).

\section{Experimental Evaluation}
\label{sec:eval}
\subsection{Experimental Setup}
\begin{table*}[t]
\centering
\caption{Complete-solve speedup relative to each original GPU program. Mean and best use all three searches. Q counts qualified deliveries. OE denotes optimization enhancement and MV denotes mandatory validation. Failed searches score $1\times$. Geomean is the geometric mean over the five projects. Bold marks the largest mean and best for each project. Section~\ref{sec:mandatory} describes the validation procedure for each project.}
\label{tab:primary}
\small
\setlength{\tabcolsep}{2pt}
\renewcommand{\arraystretch}{0.92}
\begin{tabular*}{\textwidth}{@{\extracolsep{\fill}}l *{5}{rrr}@{}}
\toprule
& \multicolumn{3}{c}{Generic-A} & \multicolumn{3}{c}{\mbox{\ours-A (OE)}} & \multicolumn{3}{c}{Generic-D} & \multicolumn{3}{c}{\mbox{\ours-D (OE)}} & \multicolumn{3}{c}{\ours-D (MV)} \\
\cmidrule(lr){2-4}\cmidrule(lr){5-7}\cmidrule(lr){8-10}\cmidrule(lr){11-13}\cmidrule(lr){14-16}
Project & Mean & Best & Q & Mean & Best & Q & Mean & Best & Q & Mean & Best & Q & Mean & Best & Q \\
\midrule
CULS & 2.581 & 3.147 & 3/3 & \textbf{3.709} & 3.906 & 3/3 & 1.303 & 1.909 & 1/3 & 2.167 & 3.216 & 2/3 & 3.020 & \textbf{4.372} & 3/3 \\
IG & 1.121 & 1.264 & 3/3 & 1.506 & 1.543 & 3/3 & 1.000 & 1.000 & 0/3 & 1.541 & \textbf{2.624} & 1/3 & \textbf{1.638} & 2.098 & 2/3 \\
DP & 1.033 & 1.043 & 3/3 & 1.213 & 1.249 & 3/3 & 1.148 & 1.221 & 2/3 & 1.085 & 1.254 & 1/3 & \textbf{1.237} & \textbf{1.334} & 3/3 \\
XP & 1.030 & 1.037 & 3/3 & \textbf{1.340} & 1.402 & 3/3 & 1.099 & 1.168 & 2/3 & 1.242 & \textbf{1.727} & 1/3 & 1.282 & 1.435 & 2/3 \\
GCS & 1.028 & 1.035 & 3/3 & 1.052 & 1.062 & 3/3 & 1.043 & 1.076 & 2/3 & 1.019 & 1.056 & 1/3 & \textbf{1.061} & \textbf{1.093} & 3/3 \\
\midrule
Geomean & 1.259 & 1.348 & & \textbf{1.570} & 1.621 & & 1.114 & 1.240 & & 1.356 & \textbf{1.807} & & 1.528 & 1.806 & \\
Total Q & & & 15/15 & & & 15/15 & & & 7/15 & & & 6/15 & & & 13/15 \\
\bottomrule
\end{tabular*}
\end{table*}

We compare generic Astra (Generic-A) and generic DeepSeek (Generic-D) with Astra under optimization enhancement [\ours-A (OE)] and DeepSeek under both modes [\ours-D (OE) and \ours-D (MV)]. Each configuration runs three searches per project. Because Generic-A already qualifies in all 15, we run Astra only with OE and leave its MV variant to future work. Astra uses \texttt{gpt-6-astra} at high reasoning effort, and DeepSeek uses \texttt{deepseek-v4.1-flash} with thinking enabled at maximum effort. Each search is capped at eight hours, 2,048 requests, and 256 million tokens. All searches and evaluations ran with CUDA 12.6 and two CPU threads.

For every configuration on all five projects, we timed each qualified nominee $P$ against $B_0$ on the same RTX 3090. On each input, we discarded one warm-up pair and timed five pairs, alternating which program ran first. For GCS, both programs used the same deterministic reduction configuration to eliminate floating-point variation from reduction order.

\subsection{Qualified Delivery and Complete-Solve Speedup}

Generic Astra already delivers a qualified program in all 15 searches (Table~\ref{tab:primary}). Generic DeepSeek delivers 7/15, yet its three-search mean exceeds generic Astra's on both placers and the timer, even with failed searches included. Of generic DeepSeek's eight failures, three involved GPU write races (two CULS, one IG), and four changed the required computation or input semantics: one IG nominee hard-coded nine layers, another replaced the full edge set with a spanning tree, an XP nominee truncated long numeric tokens, and a GCS nominee stopped numerical integration early. The eighth nominee, on DP, produced an illegal placement on \texttt{bigblue3} while $B_0$ passed the same check.

With optimization enhancement, Astra maintains 15/15 qualified deliveries and surpasses generic Astra in both mean and best speedup on every project. On IG, its mean complete-route speedup rises from 1.121$\times$ to 1.506$\times$. Even its slowest router (1.486$\times$) beats generic Astra's best (1.264$\times$).

For DeepSeek, comparing \ours-D (OE) and \ours-D (MV) shows that skill guidance alone is insufficient for a less reliable model, whereas pairing it with mandatory validation converts high-leverage edits into qualified deliveries (Tables~\ref{tab:primary} and~\ref{tab:guarded-runs}). With OE, the shared patterns help DeepSeek find faster implementations than Generic-D, raising best qualified speedup on four projects: CULS ($1.909\times$ to $3.216\times$), IG ($1.000\times$ to $2.624\times$), DP ($1.221\times$ to $1.254\times$), and XP ($1.168\times$ to $1.727\times$). With validation optional, however, its qualified delivery stays comparable to Generic-D's ($6/15$ versus $7/15$). It qualifies in only one of three searches on IG, DP, XP, and GCS, and its three-search mean falls below Generic-D's on DP and GCS.
Its nine failures stemmed from unvalidated cache reuse, GPU write races, out-of-bounds memory accesses, and legality or QoR violations exposed on the full evaluation inputs.
Enforcing MV closes this gap: qualified delivery rises from $6/15$ to $13/15$ (86.7\%), mean speedup exceeds Generic-A, Generic-D, and \ours-D (OE) on every project, best speedup exceeds both generic baselines, and the five-project geometric mean rises from $1.356\times$ to $1.528\times$. Overall, \ours-D produces the fastest qualified program on every project (MV on CULS, DP, and GCS; OE on IG and XP), while \ours-D (MV) achieves the highest mean on IG, DP, and GCS and \ours-A (OE) leads in mean on CULS and XP.

In MV mode, XP Search~3 failed the source audit: its Bookshelf reader discarded the field immediately before an adjacent \texttt{\#} comment marker, losing a net name and an auxiliary filename that $B_0$ preserved.

Beyond the patterns in Section~\ref{sec:patterns}, Astra's best XP nominee avoided token copies with views into the reader's input buffer, and its best GCS nominee transposed stage inverses for coalesced reads.

\begin{table}[t]
\centering
\caption{Complete-solve speedup ($\times$) for each DeepSeek search under optimization enhancement (OE) and mandatory validation (MV) modes. A $1.000$ entry denotes a failed search scored at the baseline fallback.}
\label{tab:guarded-runs}
\small
\setlength{\tabcolsep}{2pt}
\renewcommand{\arraystretch}{0.92}
\begin{tabular*}{\columnwidth}{@{\extracolsep{\fill}}l rrr rrr@{}}
\toprule
& \multicolumn{3}{c}{OE} & \multicolumn{3}{c}{MV} \\
\cmidrule(lr){2-4}\cmidrule(lr){5-7}
Project & S1 & S2 & S3 & S1 & S2 & S3 \\
\midrule
CULS & 2.286 & 1.000 & 3.216 & 2.404 & 2.284 & 4.372 \\
IG & 1.000 & 2.624 & 1.000 & 2.098 & 1.816 & 1.000 \\
DP & 1.254 & 1.000 & 1.000 & 1.130 & 1.334 & 1.246 \\
XP & 1.000 & 1.727 & 1.000 & 1.435 & 1.411 & 1.000 \\
GCS & 1.000 & 1.056 & 1.000 & 1.013 & 1.093 & 1.076 \\
\bottomrule
\end{tabular*}
\end{table}

\subsection{Case Study: Optimizing and Validating InstantGR}
\label{sec:validation-eval}
\textbf{Optimization enhancement.} Rerouting a net changes demand along the tracks it touches. Applying the incremental-reuse skill pattern, the agent identifies which cumulative routing costs can be reused and which must be refreshed before the next routing decision. All three Astra nominees retained cost snapshots and recomputed prefixes for affected tracks. Unlike the Generic-D router that substituted a spanning tree, they preserved the complete set of pairwise Manhattan edges used to split nets. Two nominees fused detour initialization, and the fastest fused prefix updates with demand accounting and state reset.

\textbf{Mandatory validation.} In MV Search~1, DeepSeek used the same incremental-reuse pattern and executed 61 diagnostics. Of these, 37 ran complete routes with the official checker and 24 ran native drivers. One driver compared the tree lengths and branches returned by the original and optimized FLUTE routines~\cite{flute} across ten pin counts. Another driver crashed because it omitted lookup-table initialization; DeepSeek reproduced the crash on $B_0$, then repaired and reran the driver. The submitted router qualified on all ten designs at 2.098$\times$. A second search delivered 1.816$\times$, and the third exhausted its build budget, yielding the 2/3 IG result in Table~\ref{tab:primary}. Our post-hoc review also found an out-of-domain input in a batching test, which the controller cannot detect.

\subsection{Development Cost and Limitations}
For \ours-A (OE) on IG, each search took only 17--25 minutes, including builds and tests, because Astra nominated a build as soon as it passed all four public inputs. Independently checking and timing the submitted router took about 203 minutes per nominee. Once delivered, the three Astra routers saved an average of 352\,s per run over the ten designs, measured by summing paired solve-time differences across inputs. Across all five projects, 162 of the 165 \ours-A (OE) nominee--input pairs (3 searches $\times$ 55 inputs) achieved at least $1\times$ complete-solve speedup. The three small regressions occurred on GCS (0.993--0.999$\times$).

\textbf{Limitations.} Running and evaluating the 15 searches of one configuration takes over two days, so we limited each configuration to three searches per project on one GPU model. Placement is the only task family with two solvers. The controller checks that tests ran and passed, but whether a test input is legal and discriminating depends on the agent. Extending the frozen library to more GPU EDA solvers as they become open source is future work.

\section{Conclusion}
\ours\ uses EDA knowledge to identify both the work a GPU solver can avoid and the conditions under which it can safely do so. With \ours, Astra improved mean and best complete-solve speedup on all five projects in $15/15$ qualified searches. With DeepSeek, \ours\ produced the fastest qualified program on every project, and mandatory validation raised delivery from $7/15$ ($6/15$ with OE alone) to $13/15$ while improving mean speedup on all five projects. Future work will automate skill distillation from execution traces and evaluate zero-shot skill transfer to unseen EDA tasks.

% DAC 2027 requires all references on a separate final page (at most one page).
\clearpage
\ifIEEEFormat\IEEEtriggeratref{20}\fi
\bibliography{ref}

@inproceedings{dreamplace,
  title = {{DREAMPlace: Deep Learning Toolkit-Enabled GPU Acceleration for Modern VLSI Placement}},
  author = {Lin, Yibo and Dhar, Shounak and Li, Wuxi and Ren, Haoxing and Khailany, Brucek and Pan, David Z.},
  year = {2019},
  booktitle = {Proc. DAC},
  pages = {1--6},
  doi = {10.1145/3316781.3317803}
}

@inproceedings{culs,
  title = {{Rethinking AIG Resynthesis in Parallel}},
  author = {Liu, Tianji and Young, Evangeline F.Y.},
  year = {2023},
  booktitle = {Proc. DAC},
  pages = {1--6},
  doi = {10.1109/dac56929.2023.10247961}
}

@inproceedings{instantgr,
  title = {{InstantGR: Scalable GPU Parallelization for Global Routing}},
  author = {Lin, Shiju and Xiao, Liang and Liu, Jinwei and Young, Evangeline F. Y.},
  year = {2024},
  booktitle = {Proc. ICCAD},
  pages = {1--8},
  doi = {10.1145/3676536.3676787}
}

@article{xplace,
  title = {{Xplace: An Extremely Fast and Extensible Placement Framework}},
  author = {Liu, Lixin and Fu, Bangqi and Lin, Shiju and Liu, Jinwei and Young, Evangeline F. Y. and Wong, Martin D. F.},
  year = {2024},
  journal = {IEEE Trans. Comput.-Aided Design Integr. Circuits Syst.},
  volume = {43},
  number = {6},
  pages = {1872--1885},
  doi = {10.1109/tcad.2023.3346291}
}

@inproceedings{gcstimer,
  title = {{GCS-Timer: GPU-Accelerated Current Source Model Based Static Timing Analysis}},
  author = {Lin, Shiju and Guo, Guannan and Huang, Tsung-Wei and Sheng, Weihua and Young, Evangeline and Wong, Martin},
  year = {2024},
  booktitle = {Proc. DAC},
  pages = {1--6},
  doi = {10.1145/3649329.3655983}
}

@article{chateda,
  title = {{ChatEDA: A Large Language Model Powered Autonomous Agent for EDA}},
  author = {Wu, Haoyuan and He, Zhuolun and Zhang, Xinyun and Yao, Xufeng and Zheng, Su and Zheng, Haisheng and Yu, Bei},
  year = {2024},
  journal = {IEEE Trans. Comput.-Aided Design Integr. Circuits Syst.},
  volume = {43},
  number = {10},
  pages = {3184--3197},
  doi = {10.1109/tcad.2024.3383347}
}

@inproceedings{selfabc,
  title = {{Autonomous Evolution of EDA Tools: Multi-Agent Self-Evolved ABC}},
  author = {Yu, Cunxi and Ren, Haoxing},
  year = {2026},
  booktitle = {Proc. DAC},
  pages = {1--7},
  doi = {10.1145/3770743.3804221}
}

@inproceedings{rtlfixer,
  title = {{RTLFixer: Automatically Fixing RTL Syntax Errors with Large Language Model}},
  author = {Tsai, Yunda and Liu, Mingjie and Ren, Haoxing},
  year = {2024},
  booktitle = {Proc. DAC},
  pages = {1--6},
  doi = {10.1145/3649329.3657353}
}

@inproceedings{verilogcoder,
  title = {{VerilogCoder: Autonomous Verilog Coding Agents with Graph-based Planning and Abstract Syntax Tree (AST)-based Waveform Tracing Tool}},
  author = {Ho, Chia-Tung and Ren, Haoxing and Khailany, Brucek},
  year = {2025},
  booktitle = {Proc. AAAI},
  volume = {39},
  number = {1},
  pages = {300--307},
  doi = {10.1609/aaai.v39i1.32007}
}

@inproceedings{analogcoder,
  title = {{AnalogCoder: Analog Circuit Design via Training-Free Code Generation}},
  author = {Lai, Yao and Lee, Sungyoung and Chen, Guojin and Poddar, Souradip and Hu, Mengkang and Pan, David Z. and Luo, Ping},
  year = {2025},
  booktitle = {Proc. AAAI},
  volume = {39},
  number = {1},
  pages = {379--387},
  doi = {10.1609/aaai.v39i1.32016}
}

@inproceedings{mage,
  title = {{MAGE: A Multi-Agent Engine for Automated RTL Code Generation}},
  author = {Zhao, Yujie and Zhang, Hejia and Huang, Hanxian and Yu, Zhongming and Zhao, Jishen},
  year = {2025},
  booktitle = {Proc. DAC},
  pages = {1--7},
  doi = {10.1109/dac63849.2025.11133191}
}

@inproceedings{crop,
  title = {{CROP: Circuit Retrieval and Optimization with Parameter Guidance using LLMs}},
  author = {Pan, Jingyu and Jacobson, Isaac and Zhao, Zheng and Chen, Tung-Chieh and Zhou, Guanglei and Chang, Chen-Chia and Rashingkar, Vineet and Chen, Yiran},
  year = {2025},
  booktitle = {Proc. ICCAD},
  pages = {1--9},
  doi = {10.1109/iccad66269.2025.11240643}
}

@inproceedings{kernelbench,
  title = {{KernelBench: Can LLMs Write Efficient GPU Kernels?}},
  author = {Ouyang, Anne and Guo, Simon and Arora, Simran and Zhang, Alex L and Hu, William and Re, Christopher and Mirhoseini, Azalia},
  booktitle = {Proc. International Conference on Machine Learning},
  series = {Proceedings of Machine Learning Research},
  volume = {267},
  pages = {47356--47415},
  year = {2025},
  url = {https://proceedings.mlr.press/v267/ouyang25a.html}
}

@misc{astra,
  title = {{Astra: A Multi-Agent System for GPU Kernel Performance Optimization}},
  author = {Anjiang Wei and Tianran Sun and Yogesh Seenichamy and Hang Song and Anne Ouyang and Azalia Mirhoseini and Ke Wang and Alex Aiken},
  year = {2025},
  howpublished = {arXiv:2509.07506v2},
  url = {https://arxiv.org/abs/2509.07506v2}
}

@misc{cudaforge,
  title = {{CudaForge: An Agent Framework with Hardware Feedback for CUDA Kernel Optimization}},
  author = {Zijian Zhang and Rong Wang and Shiyang Li and Yuebo Luo and Mingyi Hong and Caiwen Ding},
  year = {2025},
  howpublished = {arXiv:2511.01884v2},
  url = {https://arxiv.org/abs/2511.01884v2}
}

@misc{kernelclaw,
  title = {{KernelClaw: A Multi-Agent Framework for GPU Kernel Optimization with Dual-Memory Mechanisms}},
  author = {Qitong Sun and Jun Han and Tianlin Li and Zhe Tang and Sheng Chen and Fei Yang and Aishan Liu and Xianglong Liu and Yang Liu},
  year = {2026},
  howpublished = {arXiv:2603.10085v2},
  url = {https://arxiv.org/abs/2603.10085v2}
}

@misc{hiera,
  title = {{HIERA: Workload-Aware Planning Across Implementation Spaces for GPU Kernel Optimization}},
  author = {Jinghao Wang and Qiqi Gu and Chenpeng Wu and Jianguo Yao and Haibing Guan and Xijun Li},
  year = {2026},
  howpublished = {arXiv:2608.21157v1},
  url = {https://arxiv.org/abs/2608.21157v1}
}

@misc{audopeda,
  title = {{Automated QoR improvement in OpenROAD with coding agents}},
  author = {Amur Ghose and Junyeong Jang and Andrew B. Kahng and Jakang Lee},
  year = {2026},
  howpublished = {arXiv:2601.06268v2},
  url = {https://arxiv.org/abs/2601.06268v2}
}

@misc{pcpo,
  title = {{Self-Evolving Algorithm-Design Agents: Escaping In-Context Evolutionary Stagnation via Population-Curated Policy Optimization}},
  author = {Chen Lu and Ke Xue and Siyuan Xu and Mingxuan Yuan and Chao Qian},
  year = {2026},
  howpublished = {arXiv:2609.38757v1},
  url = {https://arxiv.org/abs/2609.38757v1}
}

@misc{autoeda,
  title = {{AutoEDA: Enabling EDA Flow Automation through Microservice-Based LLM Agents}},
  author = {Yiyi Lu and Hoi Ian Au and Junyao Zhang and Jingyu Pan and Guanglei Zhou and Yiting Wang and Jingwei Sun and Ang Li and Jianyi Zhang and Hai Li and Yiran Chen},
  year = {2026},
  howpublished = {arXiv:2508.01012v2},
  url = {https://arxiv.org/abs/2508.01012v2}
}

@misc{mcp4eda,
  title = {{MCP4EDA: LLM-Powered Model Context Protocol RTL-to-GDSII Automation with Backend Aware Synthesis Optimization}},
  author = {Yiting Wang and Wanghao Ye and Yexiao He and Yiran Chen and Gang Qu and Ang Li},
  year = {2025},
  howpublished = {arXiv:2507.19570v1},
  url = {https://arxiv.org/abs/2507.19570v1}
}

@misc{jarvis,
  title = {{JARVIS: A Multi-Agent Code Assistant for High-Quality EDA Script Generation}},
  author = {Ghasem Pasandi and Kishor Kunal and Varun Tej and Kunjal Shah and Hanfei Sun and Sumit Jain and Chunhui Li and Chenhui Deng and Teodor-Dumitru Ene and Haoxing Ren and Sreedhar Pratty},
  year = {2025},
  howpublished = {arXiv:2505.14978v2},
  url = {https://arxiv.org/abs/2505.14978v2}
}

@misc{horizon,
  title = {{Agentic Hardware Design as Repository-Level Code Evolution}},
  author = {Cunxi Yu and Chenhui Deng and Nathaniel Pinckney and Brucek Khailany},
  year = {2026},
  howpublished = {arXiv:2606.28279v1},
  url = {https://arxiv.org/abs/2606.28279v1}
}

@misc{trace2skill,
  title = {{Trace2Skill: Verifier-Guided Skill Evolution for Long-Context EDA Agents}},
  author = {Zijian Du and Nathaniel Pinckney},
  year = {2026},
  howpublished = {arXiv:2605.21810v1},
  url = {https://arxiv.org/abs/2605.21810v1}
}

@misc{phoenixbench,
  title = {{Is Agentic AI Ready for Real-World Hardware Engineering? A Deep Dive with Phoenix-bench}},
  author = {Qingyun Zou and Feng Yu and Hongshi Tan and Bingsheng He and WengFai Wong},
  year = {2026},
  howpublished = {arXiv:2605.15226v1},
  url = {https://arxiv.org/abs/2605.15226v1}
}

@misc{grevolve,
  title = {{GR-Evolve: Design-Adaptive Global Routing via LLM-Driven Algorithm Evolution}},
  author = {Taizun Jafri and Vidya A. Chhabria},
  year = {2026},
  howpublished = {arXiv:2604.22234v1},
  url = {https://arxiv.org/abs/2604.22234v1}
}

@misc{satlution,
  title = {{Autonomous Code Evolution Meets NP-Completeness}},
  author = {Cunxi Yu and Rongjian Liang and Chia-Tung Ho and Haoxing Ren},
  year = {2025},
  howpublished = {arXiv:2509.07367v1},
  url = {https://arxiv.org/abs/2509.07367v1}
}

@misc{alpharoute,
  title = {{AlphaRoute: Large Language Models as Semantic Optimizers for Multi-Objective Routing}},
  author = {Kabir Murjani and Mishri Bhavsar and Manish I. Patel and Jonti Talukdar},
  year = {2026},
  howpublished = {arXiv:2607.19768v1},
  url = {https://arxiv.org/abs/2607.19768v1}
}

@inproceedings{abc,
  title = {{ABC: An Academic Industrial-Strength Verification Tool}},
  author = {Brayton, Robert and Mishchenko, Alan},
  year = {2010},
  booktitle = {Proc. CAV},
  series = {LNCS},
  volume = {6174},
  pages = {24--40},
  doi = {10.1007/978-3-642-14295-6_5}
}

@inproceedings{epfl,
  title = {{The EPFL Combinational Benchmark Suite}},
  author = {Amar{\'u}, Luca and Gaillardon, Pierre-Emmanuel and De Micheli, Giovanni},
  year = {2015},
  booktitle = {Proc. IWLS}
}

@inproceedings{ispd2005,
  title = {{The ISPD2005 Placement Contest and Benchmark Suite}},
  author = {Nam, Gi-Joon and Alpert, Charles J. and Villarrubia, Paul and Winter, Bruce and Yildiz, Mehmet},
  year = {2005},
  booktitle = {Proc. ISPD},
  pages = {216--220},
  doi = {10.1145/1055137.1055182}
}

@inproceedings{ispd2006,
  title = {{ISPD 2006 Placement Contest}},
  author = {Nam, Gi-Joon},
  year = {2006},
  booktitle = {Proc. ISPD},
  pages = {167},
  doi = {10.1145/1123008.1123042}
}

@inproceedings{ispd2024,
  title = {{GPU/ML-Enhanced Large Scale Global Routing Contest}},
  author = {Liang, Rongjian and Agnesina, Anthony and Liu, Wen-Hao and Ren, Haoxing},
  year = {2024},
  booktitle = {Proc. ISPD},
  pages = {269--274},
  doi = {10.1145/3626184.3639693}
}

@article{rosettastone,
  title = {{RosettaStone: Connecting the Past, Present, and Future of Physical Design Research}},
  author = {Kahng, Andrew B. and Kim, Minsoo and Kim, Seungwon and Woo, Mingyu},
  year = {2022},
  journal = {IEEE Design \& Test},
  volume = {39},
  number = {5},
  pages = {70--78},
  doi = {10.1109/mdat.2022.3179247}
}

@article{asap7,
  title = {{ASAP7: A 7-nm finFET Predictive Process Design Kit}},
  author = {Clark, Lawrence T. and Vashishtha, Vinay and Shifren, Lucian and Gujja, Aditya and Sinha, Saurabh and Cline, Brian and Ramamurthy, Chandarasekaran and Yeric, Greg},
  year = {2016},
  journal = {Microelectronics Journal},
  volume = {53},
  pages = {105--115},
  doi = {10.1016/j.mejo.2016.04.006}
}

@article{flute,
  title = {{FLUTE: Fast Lookup Table Based Rectilinear Steiner Minimal Tree Algorithm for VLSI Design}},
  author = {Chu, Chris and Wong, Yiu-Chung},
  year = {2008},
  journal = {IEEE Trans. Comput.-Aided Design Integr. Circuits Syst.},
  volume = {27},
  number = {1},
  pages = {70--83},
  doi = {10.1109/TCAD.2007.907068}
}

@misc{gso,
  title = {{GSO: Challenging Software Optimization Tasks for Evaluating SWE-Agents}},
  author = {Manish Shetty and Naman Jain and Jinjian Liu and Vijay Kethanaboyina and Koushik Sen and Ion Stoica},
  year = {2025},
  howpublished = {arXiv:2505.23671v2},
  url = {https://arxiv.org/abs/2505.23671v2}
}

@misc{sweperf,
  title = {{SWE-Perf: Can Language Models Optimize Code Performance on Real-World Repositories?}},
  author = {Xinyi He and Qian Liu and Mingzhe Du and Lin Yan and Zhijie Fan and Yiming Huang and Zejian Yuan and Zejun Ma},
  year = {2025},
  howpublished = {arXiv:2507.12415v1},
  url = {https://arxiv.org/abs/2507.12415v1}
}

\end{document}